\def\year{2020}\relax
\documentclass[letterpaper]{article} 
\usepackage{aaai20}  
\usepackage{times}  
\usepackage{helvet} 
\usepackage{courier}  
\usepackage[hyphens]{url}  
\usepackage{graphicx} 
\usepackage{graphicx}  
\usepackage{amsfonts}
\usepackage{amsmath}
\usepackage{cases}
\newtheorem{theorem}{Theorem}[section]
\newtheorem{prop}{Proposition}[section]

\title{Eigenvalue Normalized Recurrent Neural Networks for Short Term Memory}
\author{Kyle Helfrich, Qiang Ye\\
\textsuperscript{\rm }Mathematics Department, University of Kentucky\\ 
Lexington, Kentucky 40509, United States\\
\{kyle.helfrich,qye3\}@uky.edu 
}

\begin{document}

\maketitle

\begin{abstract}
Several variants of recurrent neural networks (RNNs) with orthogonal or unitary recurrent matrices have recently been developed to mitigate the vanishing/exploding gradient problem and to model long-term dependencies of sequences. However, with the eigenvalues of the recurrent matrix on the unit circle, the recurrent state retains all input information which may unnecessarily consume model capacity. In this paper, we address this issue by proposing an  architecture that expands upon an orthogonal/unitary RNN with a state that is generated by a recurrent matrix with eigenvalues in the unit disc. Any input to this state dissipates in time and is replaced with new inputs, simulating short-term memory. A gradient descent algorithm is derived for learning such a recurrent matrix. The resulting method, called the Eigenvalue Normalized RNN (ENRNN), is shown to be highly competitive in several experiments. 
\end{abstract}

\section{Introduction}
 \label{introduction}

 Recurrent neural networks (RNNs) are a type of deep neural network that are  designed to handle sequential data.  The underlying dynamical system carries temporal information from one time step to another and captures potential dependencies among the terms of a sequence. Like other deep neural networks, the weights of an RNN are learned by gradient descent. For the  input at a time step to affect the output at a later time step, the gradients must back-propagate through each step. Since a sequence can be quite long, RNNs are prone to suffer from \textit{vanishing} or \textit{exploding} gradients as described in \cite{Bengio93} and \cite{Pascanu13}.  One consequence of this well-known problem is the difficulty of the network to model input-output dependency over a large number of time steps.

There have been many different architectures that are designed to mitigate this problem.  The most popular RNN architectures such as LSTMs \cite{Hochreiter97} and GRUs \cite{Cho14}, incorporate a gating mechanism to explicitly retain or discard information.  More recently, several different RNNs have been developed to maintain either a unitary or orthogonal recurrent weight matrix such as the unitary evolution RNN (uRNN) \cite{Arjovsky16}, Full-Capacity uRNN \cite{Wisdom16}, EUNN \cite{Jing16}, oRNN \cite{Mhammedi17}, scoRNN \cite{Helfrich18,Maduranga19}, Spectral RNN \cite{Zhang18}, nnRNN \cite{Kerg19}, and EXPRNN \cite{Casado19NIPS,Casado19ICML}.  There is also work using unitary matrices in GRUs such as the GORU \cite{Jing17}.  For other work addressing the vanishing/exploding gradient problem, see \cite{Henaff16,Le15,Wolter18}.   

In spite of the promises shown in recent work, orthogonal/unitary RNNs still have some  shortcomings. While an orthogonal RNN allows propagation of information over many time steps, it has an undesirable effect that all input information may be retained in all future states. Unlike gated architectures, orthogonal RNNs lack ``forget" mechanisms \cite{Jing17} to discard unwanted information. This consumes model capacity, making it difficult to efficiently model sequences with both long and short-term dependency.  

In this paper, we expand upon the orthogonal/unitary RNN architecture by incorporating a dissipative state to model short-term dependencies.  We call this model the Eigenvalue Normalized RNN (ENRNN).  Inspired by the  work on the Spectral Normalized Generative Adversarial Network (SN-GAN)  \cite{Miyato18}, we construct a recurrent matrix with its spectral radius (i.e. the largest absolute value of the eigenvalues) less than 1 through normalizing another parametric matrix by its spectral radius. A gradient descent algorithm is also derived that maintains this spectral radius property. Any input to this state will dissipate in time with repeat multiplication by the recurrent matrix and will be replaced with new input information, emulating a short-term memory state. This state can be concatenated with another state with an orthogonal/unitary recurrent matrix to form an RNN that has a long and short-term memory component to efficiently model long sequences. The resulting architecture falls within the existing framework of the basic RNN and is shown to be highly competitive in several experiments.

\section{Background and Related Work}
\label{background_related_work}

An RNN takes an input sequence of length $\tau$, denoted by $X_{\tau} = \left( x_{1}, x_{2}, ..., x_{\tau} \right)$, and produces an output sequence $Y_{\tau} = \left(y_{1}, y_{2}, ..., y_{\tau} \right)$ where $x_{t} \in \mathbb{R}^m$ and $y_{t} \in \mathbb{R}^p$.  The basic architecture consists of an input weight matrix $U \in \mathbb{R}^{n \times m}$, recurrent weight matrix $W \in \mathbb{R}^{n \times n}$, bias vector $b \in \mathbb{R}^{n}$, output weight matrix $V \in \mathbb{R}^{p \times n}$, and output bias vector $c \in \mathbb{R}^{p}$. If $\sigma(\cdot)$ is an activation function that is applied pointwise, then the hidden state $h_t \in \mathbb{R}^n$ and output $y_t$ at time $t$ is given by
\begin{eqnarray*}
h_t = \sigma \left(Ux_t + W h_{t-1} + b\right); 
y_t = Vh_t + c
\end{eqnarray*}

A problem with RNNs is that the gradient of $y_\tau$ with respect to $h_t$ involves repeat multiplication by the recurrent matrix $W$.  If the spectral radius of $W$ is less than one, gradients vanish but if the spectral radius is greater than one, gradients explode \cite{Bengio93,Pascanu13}.  One way to mitigate this issue is to use an orthogonal/unitary recurrent weight matrix which will preserve vector norms. Early work has shown simply initializing the recurrent matrix as identity or orthogonal may improve performance \cite{Henaff16,Le15}.  Several methods have also been developed that maintain an orthogonal/unitary recurrent matrix through different parameterizations. The uRNN \cite{Arjovsky16} parameterizes $W$ by a product of some special unitary matrices. The Full-Capacity uRNN \cite{Wisdom16} optimizes $W$ along the manifold of unitary matrices. The EUNN \cite{Jing16} parameterizes $W$ as a product of Givens rotation matrices, while the oRNN \cite{Mhammedi17} uses a product of Householder reflections. The scoRNN \cite{Helfrich18,Maduranga19} parameterizes $W$ by using a skew-symmetric or skew-Hermitian matrix through a scaled Cayley transform.  The EXPRNN \cite{Casado19ICML,Casado19NIPS} uses an exponential map.  There has also been work in using recurrent matrices that are near orthogonal by constraining singular values within a small distance of 1; see \cite{Vorontsov17,Zhang18}.
These models have demonstrated that orthogonal/unitary RNNs can mitigate the vanishing/exploding gradient problem and  successfully model long sequences.

Gated networks, such as LSTM and GRU, are popular RNN architectures that use a gating mechanism to control passing of long or short-term memory. Although quite successful, the LSTM is still prone to exploding gradients and may still require gradient clipping.  \cite{Jing17} considers GRU with an orthogonal recurrent matrix. However, using an orthogonal matrix with a gated network may not have the same    benefits of passing long-term information as in an orthogonal RNN.  Multiscale RNNs \cite{Schmidhuber92,ElHihi95,Koutnik14} stack multiple layers of RNNs whose states are updated in different time scales at different layers to process short and long-term information, but the difficulty is their need to determine the boundary structures defining different layers. Hierarchical multiscale RNN \cite{Chung17} introduces a binary boundary state similar to a gate to dynamically determine the boundary structures. FS-RNN \cite{Mujika17} uses a similar approach but allows the time scale at the lower level to be finer than the native scale of the input sequences.  The nnRNN \cite{Kerg19} uses a general non-normal matrix by constraining the modulus of all eigenvalues near 1.  Compared with these methods, ENRNN uses two interacting states with different recurrent matrices to model long and short-term memory.  The ENRNN only uses a non-normal matrix for the short-term memory component and constrains the spectral radius less than 1.  With the eigenvalues of the ENRNN recurrent matrix distributed within the unit disk, the corresponding state can learn short-term dependencies at any unspecified time scale with the added simplicity of a basic RNN.

The learning algorithm of ENRNN is motivated by  SN-GAN \cite{Miyato18}. The SN-GAN normalizes the discriminator weight matrix by its spectral norm, i.e. its largest singular value.  Here, we normalize the spectral radius of the recurrent matrix. Noting that the spectral radius is bounded by any matrix norm including the spectral norm, normalization by the spectral norm is expected to make the spectral radius of the matrix much less than $1$. We emphasize the importance in our approach to constrain the eigenvalues of the recurrent matrix rather than its singular values because the eigenvalues affect the dynamical behavior of RNN but the singular values do not.  See also \cite{Bengio94}.  For example, all orthogonal matrices have singular values equal to 1, but may define very different RNNs. 

Additional work that supports the idea of modeling short-term dependencies by constraining the eigenvalues is \cite{Kerg19,Orhan19}. Their analyses use a Fisher memory matrix to show non-normal networks may carry more memory than normal networks due to transient amplification.  Even though input information is eventually diminished with spectral radius $<1$, in the short-term it may increase. This theory shows that non-normal matrices may emulate short-term memory.  This can be explained by the pseudo spectrum theory where the spectrum of a matrix is within the unit disk but the pseudo spectrum may extend outside it.  In this case, the dynamics exhibit transient amplification (short-term increase) as determined by the pseudo spectrum but asymptotic (long-term) decay as determined by the spectrum.

\section{Eigenvalue Normalized Recurrent Neural Network}
\label{ENRNN}

Although an orthogonal/unitary recurrent weight matrix can help mitigate the vanishing/exploding gradient problem and hence allow an input to affect an output over long sequences, it does not have any mechanism to discard  information that is no longer needed.  In  sequences where certain input information is only used for the states or outputs locally, the state may be consumed with such information, reducing  its capacity for carrying other information.    

In order to improve the capacity of orthogonal/unitary RNNs to capture short-term dependencies, we introduce a dissipative state.  Let $h_{t} \in  \mathbb{R}^n$ be the hidden state consisting of two components: $h_{t}^{(L)} \in \mathbb{R}^q$ that captures long-term dependencies and $h_{t}^{(S)} \in \mathbb{R}^{n-q}$ that captures short-term dependencies.  In this scheme, $q$ is considered a hyperparameter.  Now let $W^{(L)} \in \mathbb{R}^{q \times q}$ be an orthogonal matrix used as the recurrent matrix for $h_{t}^{(L)}$ that is designed to propagate information over many time steps, and $W^{(S)} \in \mathbb{R}^{(n-q) \times (n-q)}$ which has a spectral radius less than one by normalizing with the spectral radius, see Section 3.1 for details.  
If we consider a recurrent weight matrix $W \in \mathbb{R}^{n \times n}$ of the form $W = \text{diag}\left(W^{(L)}, W^{(S)} \right)$, then a forward pass of the RNN will be: 
\begin{eqnarray}
\label{diagonal_block}
\begin{cases} 
      h_{t}^{(L)} = \sigma\left(U^{(L)}x_{t} + W^{(L)}h_{t-1}^{(L)} + b^{(L)} \right)\\
      h_{t}^{(S)} = \sigma\left(U^{(S)}x_{t} + W^{(S)}h_{t-1}^{(S)} + b^{(S)}\right)\\
      y_{t} = V^{(L)}h_{t}^{(L)} + V^{(S)}h_{t}^{(S)} + c
\end{cases}
\end{eqnarray}
Since $W^{(S)}$ has a spectral radius less than 1, the effect of any input on $h^{(S)}$ will decay quickly from repeat multiplication by $W^{(S)}$  with the rate of decay controlled by the magnitude of the eigenvalues of $W^{(S)}$.  Different eigenvalues with different magnitudes will then decay at different rates, emulating different lengths of memory.  

In this model, the output $y_{t}$ is determined from a combination of  $h_{t}^{(L)}$ and $h_{t}^{(S)}$ where $h_{t}^{(S)}$ contains information of recent input data, see equation (\ref{diagonal_block}). 
In this way, short-term memory that is needed to determine $y_t$ is stored in $h_{t}^{(S)}$,  but once $y_t$ is computed, $h_{t}^{(S)}$  will be gradually  replaced by information from new inputs.  This allows $h_{t}^{(L)}$ to store and carry only long-term memory information needed for the output.

In this architecture  (\ref{diagonal_block}), the hidden states $h^{(L)}$ and $h^{(S)}$ are separate. They carry the long and short-term memory in parallel and  the short-term state is directly used to determine output. If the task is to determine a single output from a sequence  at the end of the entire sequence, then $h^{(S)}$ does not affect the output until near the end of the sequence. In this case, it may still be beneficial to have $h^{(S)}$  accumulate short-term memory but to feed it into $h^{(L)}$ to indirectly affect the final output.  This can be done by adding a coupling block  to the recurrent matrix, 
\begin{equation}
W =
\left[
\begin{array}{c|c}
W^{(L)} & W^{(C)} \\
\hline
 & W^{(S)}
\end{array}
\right]
\label{eq:enrnn_expanded}
\end{equation}
where   $W^{(C)} \in \mathbb{R}^{q \times (n-q)}$ is called a coupling matrix.
Applying the recurrent matrix in (\ref{eq:enrnn_expanded}) to a forward pass of the RNN, we obtain:
\begin{equation}
    \label{diagonal_block_expanded}
    \begin{cases}
        h_{t}^{(L)}&=\sigma\left(U^{(L)}x_{t} + W^{(L)}h_{t-1}^{(L)} + W^{(C)}h_{t-1}^{(S)} + b^{(L)} \right) \\
      h_{t}^{(S)}&=\sigma\left(U^{(S)}x_{t} + W^{(S)}h_{t-1}^{(S)} + b^{(S)}\right) \hspace{1.57cm} \text{(3)}\\ 
      y_{t}&=V^{(L)}h_{t}^{(L)} + V^{(S)}h_{t}^{(S)} + c \nonumber
    \end{cases}
\end{equation}
\setcounter{equation}{3}
In this case, $h^{(S)}$ is generated by the same recurrence as before and stores short-term information of the inputs. However, with the coupling block, $h^{(L)}$ is determined from the current input, the short-term hidden state $h^{(S)}$ and $h^{(L)}$. This interaction is similar to the update of the internal state of an LSTM. In particular,  $h^{(S)}$ can be regarded as a preprocessing of several consecutive inputs designed to extract information to be used to update the long-term memory state $h^{(L)}$.  As an example, one can think of character inputs in a language processing problem. The short-term memory state may process the character inputs to produce word or phrase  information to be used in the long-term state $h^{(L)}$ so that $h^{(L)}$ can be devoted to processing the information at a higher level.  We believe this separation of the processing of characters from the processing at a higher level of sentences or concepts will be  more effective and efficient. 

We note that since $W$ is an upper triangular matrix, the eigenvalues of $W$ consist of the eigenvalues of both $W^{(L)}$ and $W^{(S)}$ and so has a spectral radius of at most one and this coupling does not alter the spectral properties of the recurrent matrix. For this reason, we do not allow a coupling from $h^{(L)}$ to $h^{(S)}$ because the fully dense recurrent matrix would not preserve the desired spectral properties. 

To illustrate how $h^{(S)}$ can simulate a short-term memory state, we note that since $\rho\left(W^{(S)}\right) < 1$, there exists some norm $\| \cdot \|$ such that $\| W^{(S)} \| < 1$. If we assume that this holds for the 2-norm, i.e. $\| W^{(S)} \|_2 < 1$, we formulate the following theorem. 

\begin{theorem}
\label{hs_theorem}
For an RNN as defined in Equations \ref{eq:enrnn_expanded} and \ref{diagonal_block_expanded} with a ReLU nonlinearity, if $\|W^{(S)} \|_2 < 1$ then 
 \begin{align*}
     \left\lVert \frac{\partial h_{t + \tau}^{(S)}}{\partial h_{t}^{(S)}}\right\rVert \leq \left\lVert W^{(S)} \right\rVert^{\tau}
     \mbox{and}
     \left\lVert \frac{\partial h_{t+\tau}^{(S)}}{\partial x_t} \right\rVert \leq \left\lVert W^{(S)} \right\rVert_2^{\tau}\left\lVert U^{(S)} \right\rVert
 \end{align*}
 where 
 $\left\lVert \cdot \right\rVert$ is the 2-norm.  
\end{theorem}
We remark that as $\tau$ increases, the derivative bounds in Theorem \ref{hs_theorem} go to zero, indicating the dependence of $h^{(S)}_{t + \tau}$ on $h^{(S)}_t$ and $x_t$ goes to zero. 

\subsection{ENRNN Gradient Descent}
\label{ENRNN_details}
The training of  $W^{(S)}$ by gradient descent can easily lead to a matrix with spectral radius greater than 1. To maintain $W^{(S)}$ with spectral radius less than 1, we parameterize it by another matrix 
$T \in \mathbb{R}^{(n-q) \times (n-q)}$ through the normalization 
\[
    W^{(S)} =W^{(S)} (T) := \frac{T}{\rho\left( T \right) + \epsilon}
\]
for some small $\epsilon > 0$, where $\rho(T)\in \mathbb{R}$ is the spectral radius of $T$. In this way, $W^{(S)}$ has  eigenvalues with modulus less than one and the training of $W^{(S)}$ is carried out in $T$. Namely, for an RNN loss function $L=L(W^{(S)})$  in terms of $W^{(S)}$, we regard it as a function $L=L(W^{(S)}(T))$ of $T$. Instead of optimizing with respect to $W^{(S)}$, we optimize $L=L(W^{(S)}(T))$ with respect to $T$. The gradients of such a parameterization are given below with a proof included in the supplemental material.
\setcounter{prop}{1}
\begin{prop}
\label{derivative}
Let $L=L(W):\mathbb{R}^{m \times m} \to \mathbb{R}$ be some differentiable loss function for an RNN with a recurrent weight matrix $W$ and let $\frac{\partial L}{\partial W} := \left[ \frac{\partial L}{\partial W_{i,j}}\right] \in \mathbb{R}^{m \times m}$. Let $W$ be parameterized by another matrix $T\in \mathbb{R}^{m \times m}$ as $W = \frac{T}{\rho\left(T\right) + \epsilon} $, where $\rho(T) \in \mathbb{R}$ is the spectral radius of $T$ and $\epsilon >0$ is a small positive number. If $\lambda = \alpha + i\beta$ (with  $\alpha,\beta \in \mathbb{R}$) is a simple  eigenvalue of $T$ with $|\lambda| = \rho (T)$ and if $u \in \mathbb{C}^{n}$ and $v \in \mathbb{C}^{n}$ are corresponding   right and left eigenvectors, i.e. $Tu=\lambda u$ and $v^*T=\lambda v^*$, then the gradient of $L=L(T)$ as a function of $T$ is given by:
\[
    \frac{\partial L}{\partial T}=\frac{1}{\Tilde{\rho}\left( T\right)}\left[\frac{\partial L}{\partial W} - \frac{1}{\Tilde{\rho}\left(T\right)}1_{m}^{T}\left( \frac{\partial L}{\partial W} \odot W\right)1_{m}C\right]
\]
where $C = \alpha \operatorname{Re}\left( S \right) + \beta \operatorname{Im}\left(S \right)$ with $S = \frac{\overline{v}u^{T}}{v^{*}u} \in \mathbb{C}^{m \times m}$,   $1_{m} \in \mathbb{R}^{m}$ is a vector consisting of all ones, $\Tilde{\rho}\left(T\right) = \rho\left(T \right) + \epsilon$, $*$ is the conjugate transpose operator, and $\odot$ is the Hadamard product. 
\end{prop}
Note that even though  complex eigenvalues come in conjugate pairs, selecting either $\lambda$ or $\overline{\lambda}$ in Proposition \ref{derivative} will result in an identical derivative due to conjugation of  $u$ and $v$; see the supplementary material.  In addition, when $\lambda$ is a multiple eigenvalue, the computation of $S$ involves a division by 0 or a number nearly 0. This is a rare situation and can be remedied in practice. First, it is unlikely to occur as the set of matrices with multiple eigenvalues lie on a low dimensional manifold in the space of $n \times n$ matrices and has a Lebesgue measure 0. Thus the probability of a random matrix having multiple eigenvalue is zero. Second, if a multiple or nearly multiple eigenvalue occurs, we may train using usual gradient descent without eigenvalue normalization for a few steps and return to $\frac{\partial L}{\partial T}$ when the eigenvalues are separated. This situation never occurred in our experiments.

Using Proposition \ref{derivative}, an optimizer  with learning rate $\zeta$ is used to first update $T$ which is then used to update $W^{(S)}$:
\begin{eqnarray}
\label{update_steps1}
    T_{k} \leftarrow T_{k-1} - \zeta \frac{\partial L}{\partial T_{k-1}}; \;\;\;
    W_{k}^{(S)} \leftarrow \frac{T_{k}}{\rho\left( T_{k}\right) + \epsilon}
\end{eqnarray}

A naive  approach may be to simply apply gradient descent on $W^{(S)}$ and then re-normalize $W^{(S)}$ by its spectral radius.  The problem is that the computed gradients $\frac{\partial L}{\partial W^{(S)}}$ do not take into account the effects of the normalization. Thus a steepest descent step on $W^{(S)}$ will reduce the loss function, but it may not be the case after $W^{(S)}$ is re-normalized by the spectral radius. In contrast, our approach   takes a gradient descent step on $T$, which decreases the loss function with the new $W$. Namely, the steepest descent direction $\frac{\partial L}{\partial T}$ has taken the eigenvalue normalization into account.

\subsection{Complexity}
\label{complexity}

The short-term memory matrix, $W^{(S)} \in \mathbb{R}^{(n - q) \times (n - q)}$, requires the computation of the spectral radius and the associated right/left eigenvectors as outlined in Section \ref{ENRNN_details}. This is done by using the Schur decomposition of the parameter matrix $T \in \mathbb{R}^{(n - q) \times (n - q)}$ which requires a complexity of $\mathcal{O}((n-q)^3)$ \cite{Demmel97} per mini-batch training iteration.  This is comparable in complexity to models that require $\mathcal{O}(n)^{3}$ complexity to maintain an orthogonal/unitary recurrent matrix such as scoRNN.  Note that implementation of a standard RNN requires a complexity of $\mathcal{O}\left(BLn^2\right)$ where $B$ is the batch size and $L$ is the sequence length.  This additional complexity of using the Schur decomposition will be comparable to a standard RNN when $n-q \leq BL$ which is typically the case. Alternatively, the complexity can be reduced to $\mathcal{O}((n-q)^2)$ by using the power-method instead of Schur decomposition, as discussed in SN-GAN \cite{Miyato18}. In practice we found that the power method may require an uneven  number of iterations and may actually be less  efficient than Schur deomposition.   
  
\section{Other Architecture Details}
\label{initialization}
We initialize $T$ to be a random matrix with eigenvalues uniformly distributed on the complex unit disc. This is done in a way similar to \cite{Helfrich18} as 
\begin{eqnarray}
  T = \text{diag}\left(B_{1}, .., B_{ \left \lfloor {n/2} \right \rfloor}\right)
  \mbox{}
  B_j =
	\gamma_{j}\begin{bmatrix}
		\cos{t_j} & -\sin{t_j} \\
		\sin{t_j} & \cos{t_j}
	\end{bmatrix}
\end{eqnarray}
where each $t_{j}$ is sampled from $\mathcal{U}[0, \frac{\pi}{2})$ and each $\gamma_{j}$ is sampled from $\mathcal{U}[-1.0, 1.0)$. This results in eigenvalues of the form $\gamma_{j}e^{\pm i t_j}$ which are uniformly distributed on the complex unit disc. For the coupling matrix,  $W^{\left( C \right)}$, initialization is Glorot Uniform \cite{Glorot10} unless indicated otherwise.  The initial states of $h_0^{(L)}$ and $h_0^{(S)}$ are set to zero and are non-trainable.

It is unknown before hand if the largest eigenvalue should have a modulus near one, so we start by setting $W^{(S)}=T$ without eigenvalue normalization and train until  $\rho\left( T\right) > 1$, at which point eigenvalue normalization is implemented. Namely, if $\rho(T) \leq 1$, then a standard gradient descent step is taken with $W^{(S)}=T$. Once an update step results in a $\rho\left( T\right) > 1$, equation (\ref{update_steps1} is used for all subsequent training steps.

Since the ENRNN is designed to expand upon orthogonal/unitary RNNs and many of these architectures use the modReLU as defined: 
$\sigma_{\text{modReLU}}(z) = \frac{z}{|z|}\sigma_{\text{ReLU}}\left( \left| z \right| + b\right)$. We also use it on most of our experiments.

\section{Experiments}
\label{experiments}

In this section, we present four experiments to compare ENRNN with LSTM and several orthogonal/unitary RNNs.  Code for the experiments and hyperparameter settings for ENRNN are available at  \path{https://github.com/KHelfrich1/ENRNN}. We compare models using single layer networks because implementation of multi-layer networks in the literature typically involves dropout, learning rate decay, and other multi-layer hyperparameters that make comparisons difficult.  This is also the setting used in prior work on orthogonal/unitary RNNs. 
Each hidden state dimension is adjusted to match the number of trainable parameters, but ENRNN can be stacked in multiple layers. For ENRNN, the long-term recurrent matrix $W^{(L)}$ is parameterized using scoRNN \cite{Helfrich18}.  For the short-term component state, we use $\epsilon = 0$ in Proposition \ref{derivative}. Unless noted otherwise, the activation function used was modReLU. For each method, the hyperparameters tuned included the optimizer $\{$Adam, RMSProp, Adagrad$\}$, and learning rates $\{10^{-3}, 10^{-4}, 10^{-5}\}$. For scoRNN, the number of negative ones used in the parameterization of the recurrent  matrix is tuned in multiplies of $10\%$ of the hidden size. For ENRNN,  the size of the short-term state  $W^{(S)}$ is tuned in multiplies of $10\%$ of the entire hidden size up to $60\%$. For the LSTM, the forget gate bias initialization and gradient clipping threshold are tuned using integers in $\left[-4,4\right]$  and in $\left[1,10\right]$ respectively. These hyperparameters were selected using a gridsearch method.  See Table \ref{experiment_hypers} and respective sections for hyperparmeter settings.  Experiments were run using Python3, Tensorflow, and CUDA9.0 on GPUs. 

\subsection{Adding Problem}
\label{adding}
\begin{figure}[t]
    \centering
    \includegraphics[width=0.97\columnwidth]{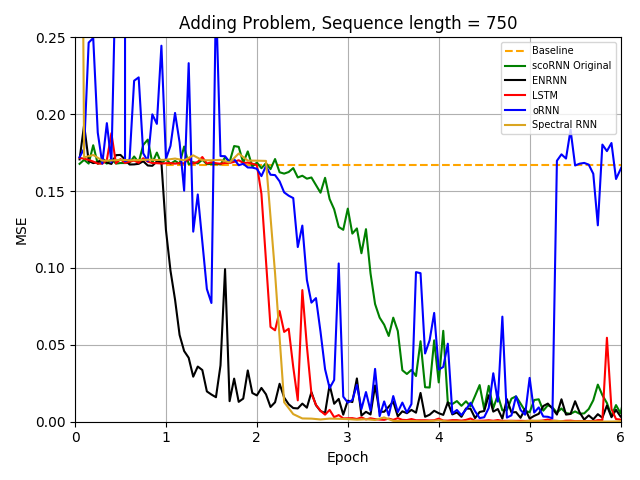}
    \caption{Test set MSE for the adding problem with sequence length of $T=750$.}
   \label{adding_fig}
\end{figure}

The adding problem  \cite{Hochreiter97} has been widely used in testing RNNs. For this experiment, we implement a variation of the adding problem \cite{Arjovsky16,Mhammedi17,Helfrich18,Maduranga19}.  The problem involves passing two sequences of length $T$ concurrently into the RNN.  The first sequence consists of entries sampled from $\mathcal{U}[0,1)$ and the second sequence consists of all zeros except for two entries that are marked by the digit one.  The first one is located uniformly in the first half of the sequence, $[1, \frac{T}{2})$ and the second one is located uniformly in the second half of the sequence, $[\frac{T}{2},T)$.  The network outputs the sum of the two entries in the first sequence that are marked by ones in the second sequence.  The loss function used is the mean squared error (MSE).  The baseline is an MSE of 0.167 which is the expected MSE for a network that predicts one regardless of the sequence. The sequence length used in this experiment was $T=750$ with training and test sets of size $100,000$ and $10,000$ examples.  
\begin{table}
	\caption{Experiment Hyperparameters. R stands for RMSProp and A for Adam optimizers.}
    \label{experiment_hypers}
	\begin{center}
	\begin{small}
    \begin{sc}
     \begin{tabular}{l c c c}
    \hline  
        Model & n & \# Params & Optimizer\\ 
        \hline \hline 
        \multicolumn{4}{c}{Adding Problem}\\
        ENRNN & 160 & $\approx 15$k & R $10^{-4}$\\
	    LSTM & 60 & $\approx 15$k & A $10^{-2}$\\
        Spectral RNN & 60 & $\approx 15$k  & A $10^{-2}$\\
        scoRNN & 170 & $\approx 15$k & R/A $10^{-4}/10^{-3}$ \\
        oRNN & 128 & $\approx 2.6$k & A $10^{-2}$\\
        Full. uRNN & 120 & $\approx 15$k & R $10^{-5}$\\
        \hline \hline
        \multicolumn{4}{c}{Copying Problem}\\
        ENRNN & 192 & $\approx 22$k & R $10^{-5}/10^{-3}$\\
	    LSTM & 68 & $\approx 22$k & R $10^{-3}$\\
	    LSTM & 192 & $\approx 158$k & R $10^{-3}$\\
        scoRNN & 190 & $\approx 22$k &R $10^{-4}/10^{-3}$ \\
        Full. uRNN & 128 & $\approx 22$k & R $10^{-3}$\\
        \hline \hline
        \multicolumn{4}{c}{TIMIT Problem}\\
        ENRNN & 468 & $\approx 200$k & A $10^{-4}$\\
	    LSTM & 158 & $\approx 200$k & R $10^{-3}$\\
	    LSTM & 468 & $\approx 1200$k & R $10^{-4}$\\
        scoRNN & 425 & $\approx 200$k & R/A $10^{-4}/10^{-3}$ \\
        \hline \hline
        \multicolumn{4}{c}{Character PTB}\\
        ENRNN & 1030 & $\approx 1016$k & A $10^{-4}/10^{-3}$\\
	    LSTM & 350 & $\approx 1016$k & R $10^{-3}$\\
	    LSTM & 1030 & $\approx 8600$k & R $10^{-3}$\\
        \hline
        \end{tabular}
    \end{sc}
    \end{small}
    \end{center}
\end{table}
The ENRNN was comprised of an $h^{(L)}$ and an $h^{(S)}$ of respective sizes $96$ and $64$ with a coupling matrix, see Equation (\ref{diagonal_block_expanded}). The $W^{(L)}$ was parameterized with $29$ negative ones.  The LSTM used an initial forget gate bias of $0$. The Spectral RNN had a learning rate decay of 0.99, r size of 16, and r margin of 0.01, similar to the settings in \cite{Zhang18}.  As per \cite{Helfrich18}, the scoRNN model used learning rates $10^{-4}$ for the recurrent weight and $10^{-3}$ for all other weights and 119 negative ones. The best hyperparameters for the oRNN were in accordance with \cite{Mhammedi17} with 16 reflections.  
Figure \ref{adding_fig} presents the convergence plots for 6 epochs. 
ENRNN converges towards 0 MSE before all other models with Spectral RNN asympototically achieving a slightly lower MSE with learning rate decay. 

\subsection{Copying Problem}
\label{copying}

The copying problem has also been used to test many orthogonal/unitary RNNs \cite{Arjovsky16,Wisdom16,Jing16,Mhammedi17,Helfrich18}. In this experiment, a sequence of digits is passed into the RNN with the first 10 digits uniformly sampled from the digits 1 through 8 followed by the marker digit 9, a sequence of $T$ zeros, and another marker digit $9$. The network is to output the first 10 digits in the sequence once it sees the second marker $9$, forcing the network to remember the original digits over the entire sequence. The total sequence length is $T+20$. The cross-entropy loss function is used. The training and test sets were $20,000$ and $1,000$ sequences, respectively. Each model was trained for $4,000$ iterations with batch size $20$. The baseline for this task is the expected cross-entropy of randomly selecting digits 1-8 after the last marker 9, $\frac{10\log(8)}{T+20}$.

\begin{figure}[t]
    \centering
   \includegraphics[width=0.97\columnwidth]{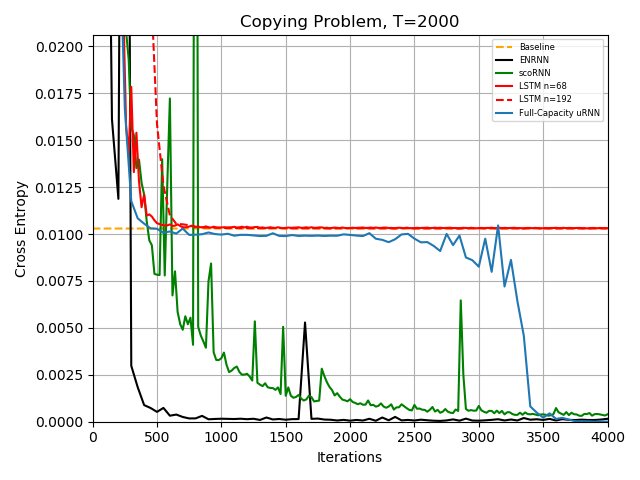}
   \caption{Cross-entropy for the copying problem with sequence length of $T =2000$.}
   \label{copying_fig}
\end{figure}

The ENRNN had an $h^{(L)}$ and $h^{(S)}$ of size $172$ and $20$ with a coupling matrix $W^{(C)}$, see Equation (\ref{diagonal_block_expanded}). A learning rate of $10^{-5}$ for $h^{(L)}$ and learning rate of $10^{-3}$ for all other weights was used. The $W^{(L)}$ was parameterized with $52$  negative ones.  The LSTM used an initial forget gate bias of $1.0$ for $n=68$ and $-2$ for $n=192$.  As per \cite{Helfrich18}, the scoRNN model used learning rate $10^{-4}$ for the recurrent weights with 95 negative ones and $10^{-3}$ for all other weights. Figure \ref{copying_fig} plots cross-entropy values for 4000 iterations. As a reference, the LSTM was also run with the same hidden size of ENRNN, $n=192$, which has $\approx 7$ times more trainable parameters than ENRNN and is still unable to drop below the baseline.  Again, ENRNN outperforms other methods.

\subsection{TIMIT}
\label{timit}

The TIMIT dataset \cite{Garofolo93}  consists of spoken sentences from 630 different speakers with eight major dialects of American English.  We used the same setup as outlined in \cite{Wisdom16}.  The data set consisted of 3,696 training, 192 testing, and a validation set of 400 audio files.  Each audio file was downsampled to 8kHz and a short-time Fourier transform (STFT) was applied  \cite{Wisdom16}.  The sequence of the log magnitude of the STFT values are fed into RNNs and the output is the same sequence shifted forward in time by 1 to predict the next log magnitude STFT value in the sequence. Each sequence was padded with zeros to make uniform lengths. The loss function was the mean squared error (MSE) and was computed by taking the squared difference between the predicted and actual log magnitudes and applying a mask to zero out padded entries before computing the batch mean.  The batch size was 28.    

Following \cite{Wisdom16}, the models were also analyzed using time-domain metrics consisting of the Signal-to-Noise Ratio (SegSNR), Perceptual Evaluation of Speech Quality (PESQ), and Short-Time Objective Intelligibility (STOI). For SegSNR, the higher the positive value indicates more signal than noise. The PESQ values range within $\left[1.0, 4.5\right]$ with a higher value indicating better signal quality. The STOI values range within $\left[0,1\right]$ with a higher value indicating better human intelligibility. To compute the scores, the predicted log-magnitudes on the test set were used to reconstruct the sound waves and were compared with the original sound waves, see \cite{Wisdom16}.  

\begin{figure}[t]
    \centering
	\includegraphics[width=0.95\columnwidth]{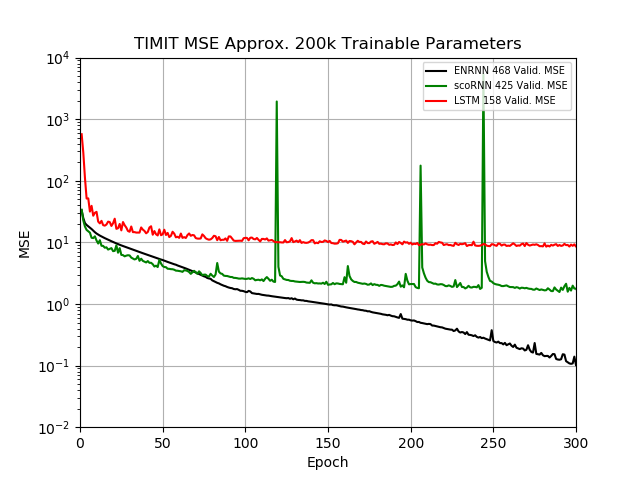}
	\caption{Validation set MSE for the TIMIT problem}
	\label{TIMIT_figure}
\end{figure}

The ENRNN consisted of $h^{(L)}$ and $h^{(S)}$ of sizes $374$ and $94$ respectively with no coupling matrix.  The number of negative ones for $W^{(L)}$ was $374$. The LSTM used gradient clipping of $1.0$ and forget gate bias initialization of $-4.0$. As per \cite{Helfrich18}, the scoRNN used learning rate $10^{-4}$ to update the recurrent matrix and learning rate $10^{-3}$ for all other weights. The number of negative ones for the recurrent weight was $43$.
Each model was trained for 300 epochs. Table \ref{timittable} reports the results of the best epoch in validation MSE scores and Figure \ref{TIMIT_figure} plots convergence of these scores. As a secondary measure, we also show in Table \ref{timittable} scores of three perceptual metrics.  As a reference, the LSTM was also run with the same hidden size of ENRNN, $n=468$, which has $\approx 6$ times more trainable parameters than ENRNN and achieves worse scores except for PESQ where it is the same.  Again, ENRNN significantly outperforms scoRNN and LSTM in the validation and testing MSEs and produces the overall best scores in the perceptual metrics.

 \begin{table}
  		\caption{TIMIT: Best  validation MSE after 300 epochs with test MSE and perceptual metrics. $N$ - dimension of $h$ (for ENRNN, dimensions of $h^{(L)}$/$h^{(S)}$) that match $\approx 200$k trainable parameters.}
          \label{timittable}
  		\begin{center}
  		\begin{small}
          \begin{sc}
          \begin{tabular}{ l  c c c  c }
       \hline
       Model & n & \#Params & Valid. & Test.\\ 
             &   &          & MSE    & MSE\\
       \hline \hline 
       ENRNN & 374/94 & $\approx 200$k &  \textbf{0.13} & \textbf{0.13}\\
       scoRNN & 425 & $\approx 200$k & 1.56 & 1.52\\
       LSTM & 158 & $\approx 200$k & 8.53 & 8.27\\
       \hline
       LSTM & 468 & $\approx 1200$k & 5.60 & 5.42\\
       \hline \hline
       Model & N & SegSNR (dB) & STOI & PESQ\\
       \hline
       ENRNN & 374/94 & \textbf{4.84} & \textbf{0.83} & \textbf{2.75}\\
       scoRNN & 425 & 4.55 & 0.82 & 2.72\\
       LSTM & 158 & 4.00 & 0.79 & 2.51\\
       \hline
       LSTM & 468 & 4.82 & 0.81 & 2.75 \\
       \end{tabular}
      \end{sc}
  \end{small}
      \end{center}
 \end{table}

\subsection{Character PTB}
\label{charPTB}

The models were also tested on a character prediction task using the Penn Treebank Corpus \cite{Marcus93}. The dataset consists of a word vocabulary of $10$k with all other words marked as $<$unk$>$, resulting in a total of $50$ characters with the training, validation, and test sets consisting of approximately $5102$k, $400$k, and $450$k respective characters.  The batch size was set to $32$.  Due to the length of each sequence, the sequences were unrolled in length of $50$ steps. Each sequence is fed into RNNs and the output is the same sequence shifted forward by one step to predict the next character.  At the end of training of each sequence in a batch, the final hidden state is passed onto the next training sequence as the initial state. We use a linear embedding layer that maps each input character to $R^N$ ($N$ being the hidden state dimension).  The loss function was cross-entropy. We report the customary performance metrics of  bits-per-character (bpc) which is the cross-entropy loss with the natural logarithm replaced by the base 2 logarithm.

ENRNN has an $h^{(L)}$ and $h^{(S)}$ of sizes $310$ and $720$ with a coupling matrix $W^{(C)}$ using a truncated orthogonal initialization, a fixed input weight matrix set to identity, and ReLU nonlinearity. Learning rate of $10^{-4}$ is used to update $W^{(L)}$ (with 186 neg. ones) and $10^{-3}$ for all other weights. The LSTM uses a forget gate bias initialization of 0.0 and gradient clipping of $8$.
We report the best results after 20 epochs training in Table \ref{charptbtable}. Also included in the table are the results from \cite{Mhammedi17,Jing17} for the same problem.  As a reference, the LSTM was also run with the same hidden size of ENRNN, $n=1030$, which results in a better score but requires $\approx 8.5$ times more trainable parameters than ENRNN.  We see that ENRNN slightly outperforms LSTM when matching the number of trainable parameters and all other models.


\begin{table}
	\caption{Character PTB: Best testing MSE in BPC after 20 epochs. $N$ - dimension of $h$ (for ENRNN, dimensions of $h^{(L)}$/$h^{(S)}$). Entries marked by an asterix (*), (**), and (***) are reported from \cite{Jing17}, \cite{Mhammedi17}, and \cite{Kerg19}, resp.}
    \label{charptbtable}
	\begin{center}
	\begin{small}
    \begin{sc}
     \begin{tabular}{ l c c c c}
    \hline  
        Model & n & \# Param & Valid. & Test\\
              &   &          & BPC    & BPC\\
        \hline \hline 
        ENRNN & 310/720 & $\approx 1016$k & \textbf{1.475} & \textbf{1.429}\\
	    LSTM        & 350 & $\approx 1016$k & 1.506 &1.461\\
        GRU       & 415 & - & -  & 1.601*\\
        EURNN      & 2048 & - & - & 1.715* \\
        GORU       & 512 & - & -  & 1.623*\\
        oRNN       & 512 & $\approx 183$k & 1.73** & 1.68**\\
        nnRNN      & 1024 & $\approx 1320$k & - & 1.47***\\
        \hline
        LSTM        & 1030 & $\approx 8600$k & 1.447 & 1.408\\
        \hline
        \end{tabular}
    \end{sc}
    \end{small}
    \end{center}
\end{table}
  
\section{Exploratory Experiments}
\label{exploratory}

We present exploratory studies to demonstrate short-term and long-term dependency of the states $h^{(S)}$ and $h^{(L)}$ respectively as intended. 

\subsection{Gradients}
\label{exploratory_gradients}

\begin{figure}[t]
   \includegraphics[width=0.96\linewidth]{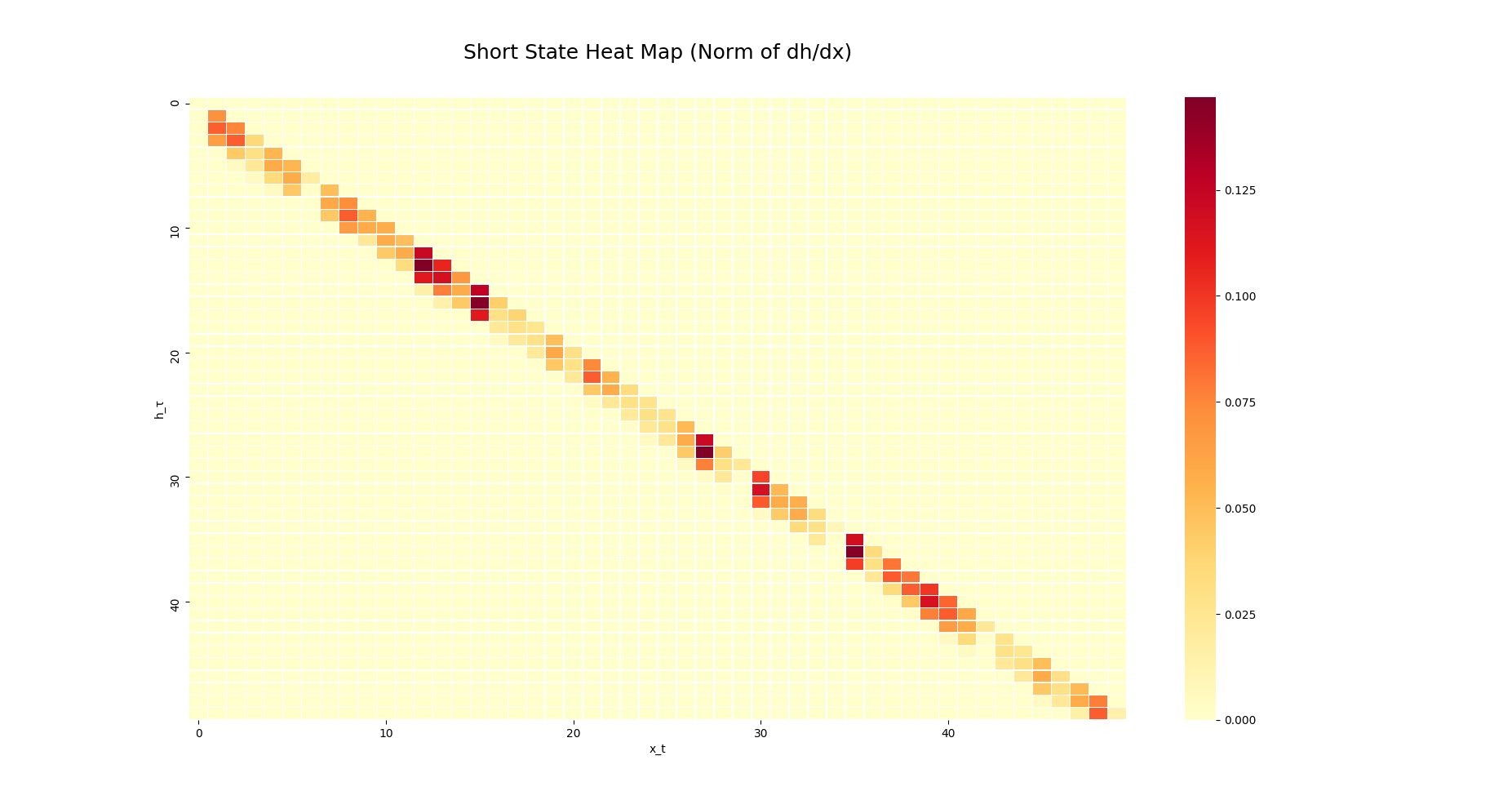}
 \caption{Gradient norms
 $\|\frac{\partial h_\tau^{(S)}}{\partial x_t}\|$  The $x$-axis is  $t$ from left to right and y-axis is $\tau$ from top to bottom. The column at $t$ shows dependence of states $h_\tau^{(S)}$/$h_\tau^{(L)}$ on $x_t$. }
 \label{heatmap1}
 \end{figure}

\begin{figure}[t]
   \includegraphics[width=0.96\linewidth]{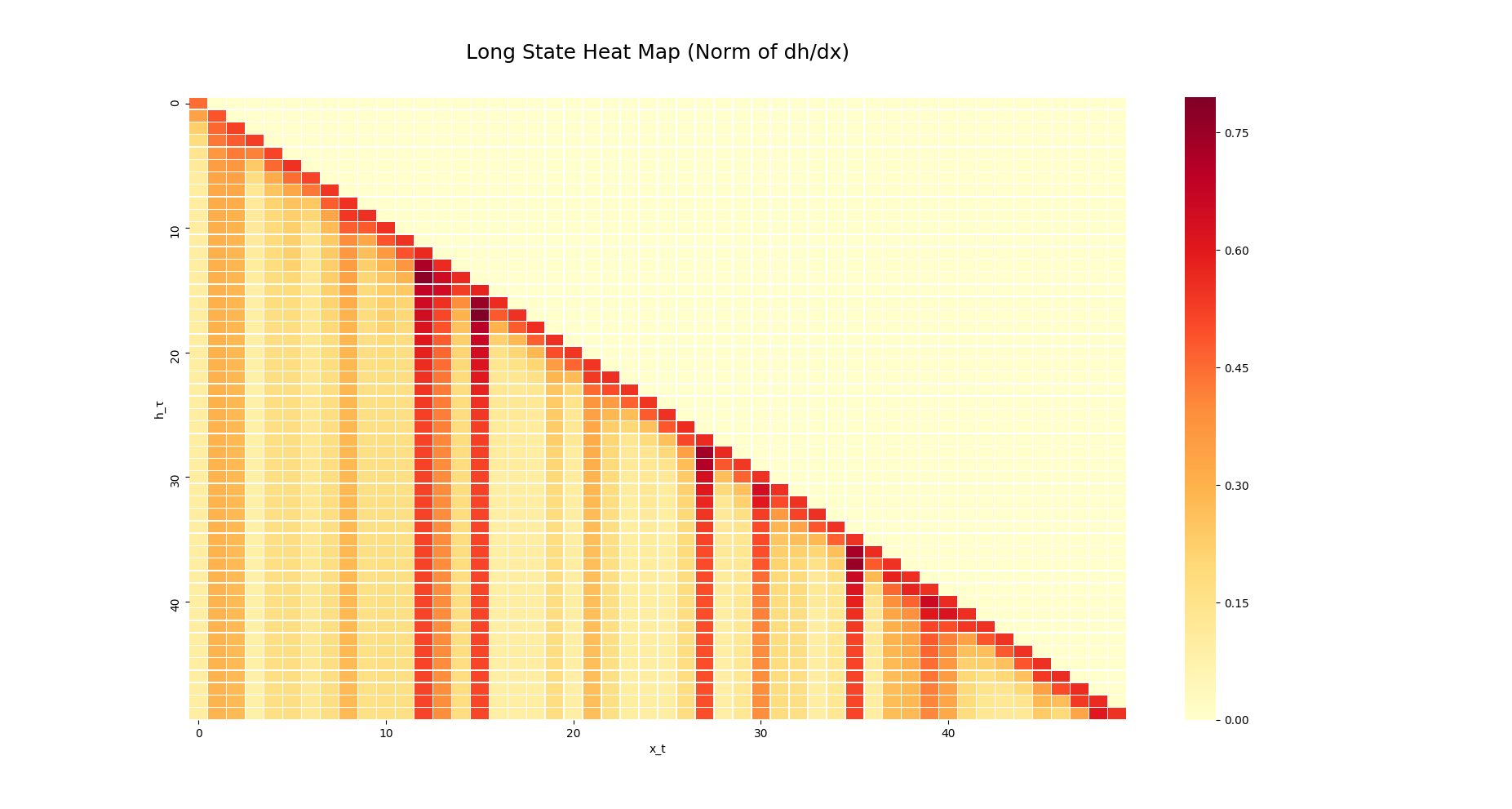}
\caption{Gradient norms
 $\|\frac{\partial h_\tau^{(L)}}{\partial x_t}\|$  
The $x$-axis is  $t$ from left to right and y-axis is $\tau$ from top to bottom. The column at $t$ shows dependence of states $h_\tau^{(S)}$/$h_\tau^{(L)}$ on $x_t$. }
 \label{heatmap2}
 \end{figure}

For each pair of time steps $t\le \tau$, we use $\lvert\lvert \frac{\partial h_{\tau}^{(S)}}{\partial x_{t}}\rvert\rvert_{2}$ and $\lvert\lvert \frac{\partial h_{\tau}^{(L)}}{\partial x_{t}}\rvert\rvert_{2}$ to measure the dependency of $h_{\tau}^{(S)}$ and $ h_{\tau}^{(L)}$ at time $\tau$ respectively on the  input $x_t$ at time $t$. We consider a small Adding Problem (Sec. 5.1) of sequence length $T=50$ using an ENRNN of hidden size $n = 40$ with $h^{(L)}$ block size of $24$ and $h^{(S)}$ block size of $16$. 
We compute the gradient norms over the first random mini-batch at the beginning of the sixth epoch and plot them as a heat map in Figure \ref{heatmap1} and \ref{heatmap2}.  Here the x-axis is the input time step (going from left to right) and the y-axis is the hidden state  time step (going from top to bottom).  

As can be seen,  the short-term state gradient (left) diminishes quickly as $\tau$ increases from $t$, demonstrating the short-term dependency of $h_{\tau}^{(S)}$. On the other hand, the long-term state gradient (right) may stay large for all $\tau$ showing long-term dependency of $h_{\tau}^{(S)}$. Of particular note, there appears to be a few vertical lines that have higher gradient norms relative to other input steps.  It appears that these inputs have a greater effect on the model than others.    

\subsection{Hidden State Sizes}
\label{hidden_state_sizes_exploratory}

\begin{figure}[t]
   \includegraphics[width=0.95\linewidth]{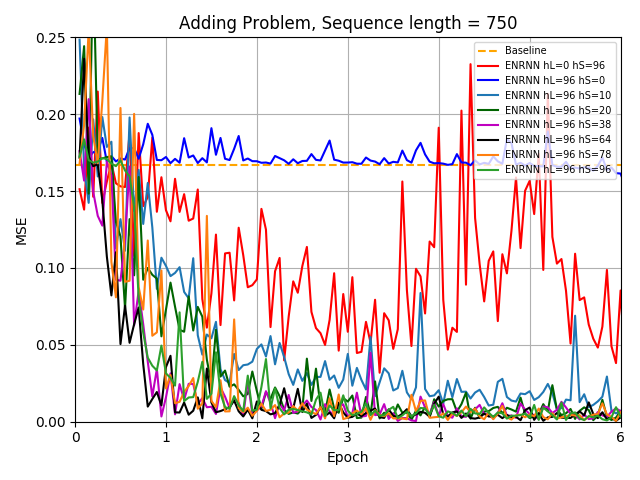}
 \caption{Test set MSE for the ENRNN on the adding problem with sequence length of $T=750$ with various short-term hidden state sizes $h^{(S)}$. }
 \label{various_hs}
 \end{figure}

\begin{figure}[t]
   \includegraphics[width=0.95\linewidth]{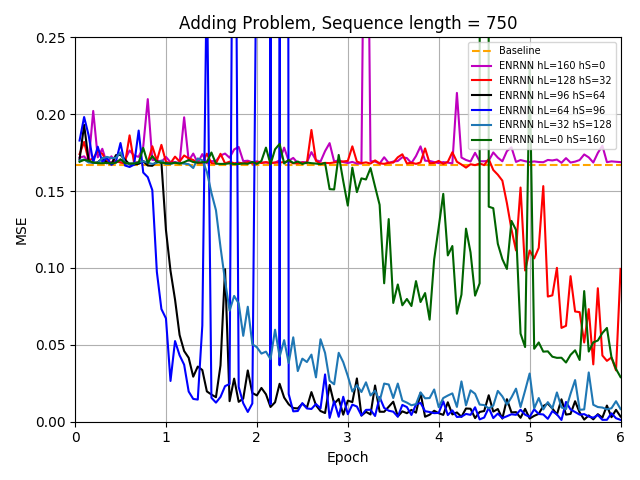}
 \caption{Test set MSE for the ENRNN on the adding problem with sequence length of $T=750$ with fixed hidden state size of $160$ and various short-term and long-term hidden state sizes $h^{(S)}$ and $h^{(L)}$. }
 \label{various_hs2}
 \end{figure}

In this section, we explore the effect of different short-term hidden states on model performance on the adding problem using similar 
settings as discussed in Section \ref{adding}. 
In Figure \ref{various_hs}, we keep the $h^{(L)}$ state size fixed at $n=96$ and adjust the $h^{(S)}$ state size from $0$ to $96$ for testing sensitivity. In Figure \ref{various_hs2}, we keep the total hidden state size fixed at $n=160$ and adjust the $h^{(L)}$ and $h^{(S)}$ sizes. In addition, we run the experiment with no $h^{(L)}$ and no $h^{(S)}$. As can be seen, having no long-term memory state, $h^{(L)}=0$, the ENRNN is unable to approach zero MSE and having no short-term memory state, $h^{(S)} = 0$, the ENRNN is only able to pass the baseline after almost 6 epochs, if at all. In general, as the size of $h^{(S)}$ increases, the performance increases with optimal performance occurring around $h^{(S)}=64$ and $h^{(L)}=96$ or $h^{(S)}=96$ and $h^{(L)} = 64$. It should be noted that for the large range $h^{(S)}>38$, near optimal performance is achieved in Figure \ref{various_hs}. 

\section{Conclusion}
\label{conclusion}

We have introduced a new RNN architecture with two components to accumulate long and short-term memory information. We have developed a gradient descent algorithm for learning a short-term recurrent matrix with eigenvalues on the unit disc.  Our exploratory study indicates that the long-term and short-term  component states behave as intended. Experimental results in four widely used examples indicate that the ENRNN is competitive with 
orthogonal/unitary RNNs and LSTM. Also important is that our method is entirely based on the basic RNN framework without the need of highly complex architectures that are more difficult to understand and implement.

\subsubsection{Acknowledgments.}
This research was supported in part by NSF under grants  DMS-1821144 and
DMS-1620082.

\bibliographystyle{aaai}
\bibliography{main}

\newpage

\section{Supplementary Material}

\subsection{Proof of Theorem 3.1}
\setcounter{section}{3}
\begin{theorem}
\label{hs_theorem}
For an RNN as defined in Equations 2 and 3 with a ReLU nonlinearity, if $\|W^{(S)} \|_2 < 1$ then 
 \begin{align*}
     \left\lVert \frac{\partial h_{t + \tau}^{(S)}}{\partial h_{t}^{(S)}}\right\rVert \leq \left\lVert W^{(S)} \right\rVert^{\tau}
     \mbox{and}
     \left\lVert \frac{\partial h_{t+\tau}^{(S)}}{\partial x_t} \right\rVert \leq \left\lVert W^{(S)} \right\rVert_2^{\tau}\left\lVert U^{(S)} \right\rVert
 \end{align*}
 where $\left\lVert \cdot \right\rVert$ is the 2-norm. 
\end{theorem}

\textbf{Proof:}
By the chain rule, we obtain:
\begin{align*}
     \frac{\partial h_{t + \tau}^{(S)}}{\partial h_{t}^{(S)}} =& \frac{\partial h^{(S)}_{t+\tau}}{\partial h^{(S)}_{t + \tau - 1}}\frac{\partial h^{(S)}_{t+\tau -1}}{\partial h^{(S)}_{t+\tau - 2}} ... \frac{\partial h^{(S)}_{t+1}}{\partial h^{(S)}_{t}}\\
     =& \prod_{k=t + \tau}^{t+1}\frac{\partial h^{(S)}_{k}}{\partial h_{k-1}^{(S)}}\\
     =& \prod_{k=t+\tau}^{t + 1}G_{k}W^{(S)}
 \end{align*}
where $G_{k} = \text{diag}\left(\sigma'(a_j)\right)$ is a diagonal matrix consisting of the derivative of the nonlinearity function for each activation $a_j$ at time step $k$. Now taking the two-norm to both sides,
\begin{align*}
    \left\lVert \frac{\partial h_{t+\tau}^{(S)}}{\partial h_{t}^{(S)}} \right\rVert_2 \leq \prod_{k=t + \tau}^{t+1}\left\lVert G_{k}\right\rVert_2 \left\lVert W^{(S)} \right\rVert_2 \leq \left\lVert W^{(S)} \right\rVert_2^{\tau}
\end{align*}
Similarly, we have
\begin{align*}
    \frac{\partial h_{t+\tau}^{(S)}}{\partial x_{t}} = \frac{\partial h_{t+\tau}^{(S)}}{dh_{t}^{(S)}}\frac{\partial h_{t}^{(S)}}{\partial x_t} = \frac{\partial h_{t+\tau}^{(S)}}{dh_{t}^{(S)}}G_{t}U^{(S)}
\end{align*}
and taking the two-norm to both sides we obtain:
\begin{align*}
    \left\lVert \frac{\partial h_{t+\tau}^{(S)}}{\partial x_{t}} \right\rVert_2 \leq \left\lVert W^{(S)} \right\rVert_2^{\tau}\left\lVert U^{(S)} \right\rVert_2
\end{align*}
$\Box$

\setcounter{section}{8}
\subsection{Proof of Proposition 3.2}
\setcounter{section}{3}
\setcounter{prop}{1}

\begin{prop}
\label{derivative}
 Let $L=L(W):\mathbb{R}^{m \times m} \to \mathbb{R}$ be some differentiable loss function for an RNN with a recurrent weight matrix $W$ and let $\frac{\partial L}{\partial W} := \left[ \frac{\partial L}{\partial W_{i,j}}\right] \in \mathbb{R}^{m \times m}$. 
 Let $W$ be parameterized by another matrix $T\in \mathbb{R}^{m \times m}$ as $W = \frac{T}{\rho\left(T\right) + \epsilon} $, where $\rho(T) \in \mathbb{R}$ is the spectral radius of $T$ and $\epsilon >0$ is a small positive number. If $\lambda = \alpha + i\beta$ (with  $\alpha,\beta \in \mathbb{R}$) is a simple  eigenvalue of $T$ with $|\lambda| = \rho (T)$ and if $u \in \mathbb{C}^{n}$ and $v \in \mathbb{C}^{n}$ are corresponding   right and left eigenvectors, i.e. $Tu=\lambda u$ and $v^*T=\lambda v^*$, then the gradient of $L=L(T)$ as a function of $T$ is given by:
 \[
    \frac{\partial L}{\partial T}=\frac{1}{\Tilde{\rho}\left( T\right)}\left[\frac{\partial L}{\partial W} - \frac{1}{\Tilde{\rho}\left(T\right)}1_{m}^{T}\left( \frac{\partial L}{\partial W} \odot W\right)1_{m}C\right]
\]
where $C = \alpha \operatorname{Re}\left( S \right) + \beta \operatorname{Im}\left(S \right)$ with $S = \frac{\overline{v}u^{T}}{v^{*}u} \in \mathbb{C}^{m \times m}$,   $1_{m} \in \mathbb{R}^{m}$ is a vector consisting of all ones, $\Tilde{\rho}\left(T\right) = \rho\left(T \right) + \epsilon$, $*$ is the conjugate transpose operator, and $\odot$ is the Hadamard product. 
\end{prop}

 \textbf{Proof:} To find the derivative with respect to the $(i,j)$ entry of $T $, we obtain by the chain rule:
 \begin{flalign}
 \label{pfline1}
 \nonumber
  \frac{\partial L}{\partial T_{i,j}} =& \sum_{k,l = 1}^{m} \frac{\partial L}{\partial W_{k,l}}\frac{\partial W_{k,l}}{\partial T_{i,j}} \\
   =& \sum_{k,l = 1}^{m} \frac{\partial L}{\partial W_{k,l}} \left(\frac{1}{\rho(W) + \epsilon}\frac{\partial T_{k,l}}{\partial T_{i,j}} + T_{k,l} \frac{\partial \left( r^{-1} \right)}{\partial T_{i,j}}\right)
 \end{flalign}
 where $r = \rho\left(T\right) + \epsilon$.
 Looking at the second term in equation (\ref{pfline1}) and using $\rho\left( T \right) = \left(\alpha^2 + \beta^2 \right)^{\frac{1}{2}}$, we have
 \begin{flalign}
 \label{dfw_rhs}
 \frac{\partial \left( r^{-1} \right)}{\partial T_{i,j}} = - \frac{1}{\rho\left( T \right)\left(\rho(T) + \epsilon \right)^{2}}\left( \alpha \frac{\partial \alpha}{\partial T_{i,j}} + \beta \frac{\partial \beta}{\partial T_{i,j}} \right)
 \end{flalign}

 Since $\lambda$ is a simple eigenvalue, we have  
 \[
 \frac{\partial \lambda}{\partial T_{ij}} = \frac{v^* \frac{\partial T}{\partial T_{ij}} u}{v^{*}u}
  = \frac{\overline{v}_i u_j }{v^{*}u} 
 \]
 and hence $ \frac{\partial \lambda}{\partial T} = \frac{\overline{v}u^{T}}{v^{*}u} = S$. So
 \begin{eqnarray}
     \label{dalpha_dbeta}
     \frac{\partial \alpha}{\partial T }= 
     \operatorname{Re}\left( S \right);  \;\;
     \frac{\partial \beta}{\partial T }=     \operatorname{Im}\left( S  \right)
 \end{eqnarray}
 Combining equations (\ref{pfline1}), (\ref{dfw_rhs}), and (\ref{dalpha_dbeta}) with the fact that $\frac{\partial W_{k,l}}{\partial W_{i,j}} = 1$ for $(k,l) = (i,j)$ and 0 otherwise, we obtain
\begin{eqnarray*}
  \nonumber
\frac{\partial L}{\partial T_{i,j}} =& \frac{1}{\Tilde{\rho\left( T \right)}}\frac{\partial L}{\partial W_{i,j}}- \frac{1}{\rho \left( T \right)\Tilde{\rho \left( T\right)^2}}\sum_{k,l=1}^{m}\frac{\partial L}{\partial W_{k,l}}T_{k,l}C_{i,j}\\
  =& \frac{1}{\Tilde{\rho\left( T \right)}}\left[\frac{\partial L}{\partial W_{i,j}} - \frac{1}{\Tilde{\rho\left(T\right)}}1_{m}^{T}\left( \frac{\partial L}{\partial W} \odot W\right)1_{m}C_{i,j} \right]
\end{eqnarray*}
  as desired.
  $\Box$

  We remark that if $W$ is a real matrix, then complex eigenvalues appear in conjugate pairs, both of which give the spectral radius $\rho (T)$. However, the formula in the above theorem is independent of which eigenvalue we use. Specifically, if $\lambda$ in the theorem is a  complex eigenvalue, $\overline{\lambda}$ is also an eigenvalue with $\overline{u}$ and $\overline{v}$ as right and left eigenvectors. Using $\overline{\lambda}$, $\overline{u}$ and $\overline{v}$ in the theorem, we obtain the same formula for 
  $\frac{\partial L}{\partial T}$ because $ \frac{{v}\overline{u}^{T}}{\overline{v}^{*}\overline{u}} =\overline{S} $ and  correspondingly 
  $ \alpha \operatorname{Re}\left( \overline{S} \right) + \beta \operatorname{Im}\left(\overline{S} \right) =C$.

\end{document}